\documentclass[final]{cvpr}

\usepackage{times}
\usepackage{epsfig}
\usepackage{graphicx}
\usepackage{amsmath}
\usepackage{amssymb}

\usepackage{bm}
\usepackage{amsfonts}
\usepackage{amstext}
\usepackage{multirow}
\usepackage[american]{babel}
\usepackage{microtype}
\usepackage{booktabs}
\usepackage{subfigure}
\usepackage{rotating}
\usepackage{pifont}
\usepackage[ruled, linesnumbered]{algorithm2e}

\newcommand{\cmark}{\ding{51}}%
\newcommand{\xmark}{\ding{55}}%

\usepackage[pagebackref=false,breaklinks=true,colorlinks,bookmarks=false]{hyperref}

\def\cvprPaperID{5229} 
\def\confYear{CVPR 2021}

\begin{document}

\title{Source-Free Domain Adaptation for Semantic Segmentation}

\renewcommand{\thefootnote}{\fnsymbol{footnote}}
\author {
  Yuang Liu~,\quad Wei Zhang\footnotemark[1]~,\quad Jun Wang\footnotemark[1] \\
  East China Normal University, Shanghai, China \\
  {\tt\small \{frankliu624, zhangwei.thu2011, wongjun\}@gmail.com}
}

\maketitle

\footnotetext[1]{Corresponding author.}
\renewcommand{\thefootnote}{\arabic{footnote}}

\begin{abstract}
  Unsupervised Domain Adaptation (UDA) can tackle the challenge that convolutional neural network~(CNN)-based approaches for semantic segmentation heavily rely on the pixel-level annotated data, which is labor-intensive. 
  However, existing UDA approaches in this regard inevitably require the full access to source datasets to reduce the gap between the source and target domains during model adaptation, which are impractical in the real scenarios where the source datasets are private, and thus cannot be released along with the well-trained source models.
  To cope with this issue, we propose a source-free domain adaptation framework for semantic segmentation, namely SFDA, in which only a well-trained source model and an unlabeled target domain dataset are available for adaptation. 
  SFDA not only enables to recover and preserve the source domain knowledge from the source model via knowledge transfer during model adaptation, but also distills valuable information from the target domain for self-supervised learning.
  The pixel- and patch-level optimization objectives tailored for semantic segmentation are seamlessly integrated in the framework.
  The extensive experimental results on numerous benchmark datasets highlight the effectiveness of our framework against the existing UDA approaches relying on source data. 
\end{abstract}

\vspace{-3mm}
\section{Introduction}

Semantic segmentation has been a critical computer vision task, which aims to segment and parse a scene image into different image regions associated with semantic categories. It is critical for precisely understanding the visual scene and can be applied to numerous potential applications, such as autonomous driving~\cite{chen2018road}, visual grounding~\cite{hu2016segmentation,xiao2017weakly,sigurdsson2020visual}, and image editing~\cite{morel2012fourier}. 
But the success of current segmentation techniques depends on large-scale densely-labeled datasets that are prohibitively expensive to be collected in reality. For instance, it takes about 90 minutes to manually annotate a Cityscapes image. An intuitive method to address this issue is transferring knowledge from existing models trained on source datasets to the unlabeled target domain. However, it tends to be hindered by the issue of domain shift which is caused by various data distributions in source and target domains. 

Unsupervised domain adaptation (UDA)~\cite{ganin2016domain,CycleGAN2017,hong2018conditional,chen2019maxsquare} for semantic segmentation has been proposed to address this issue and generalize the well-trained models on an unlabeled target domain, avoiding expensive data annotation.
All the methods suppose that both the well-trained source models and labeled source datasets are available. 
This is because source data plays a vital role in retaining valuable source knowledge during adaptation training and reducing the cross-domain discrepancy iteratively.
However, in some crucial areas like autonomous driving, the source datasets may be private and commercial, making only the source models and unlabeled target datasets available. Due to the lack of supervision of the source domain and the uncertainty of target pseudo-labels, none of these UDA methods can work in such source-free scenarios. 

\begin{figure}[t]
  \centering
  \includegraphics[width=0.92\linewidth]{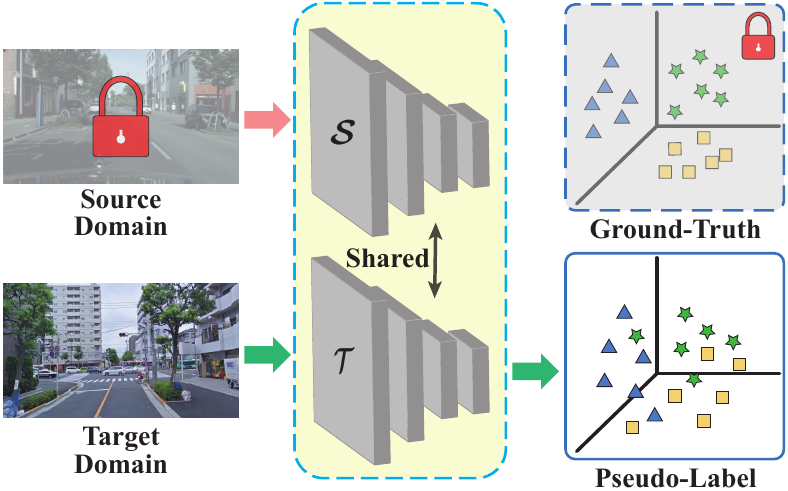}
  \caption{Overview of source-free UDA for segmentation.}
  \label{fig:overview}
  \vspace{-3mm}
\end{figure}

With these insights, we formulate a new but important problem --- source-free domain adaptation for semantic segmentation, in which only a well-trained source model and an unlabeled target domain dataset are available for adaptation. 
Recently, a tiny number of source-free UDA methods~\cite{li2020model,kundu2020universal,liang2020we,sahoo2020unsupervised,kim2020domain,liang2019distant} have been developed to tackle a similar issue on image classification.
However, the image-level computer vision task just associates the label with a whole image, which is fundamentally different from image segmentation that belongs to a pixel-level task with each pixel associated with a semantic label.
As shown in Figure~\ref{fig:overview}, the pseudo-labels of one target image contains multiple classes shifting on diverse distributions.
As such, it is nontrivial for the above methods to leverage clustering for each class adaptation.
Since considering that the source domain knowledge cannot be preserved and utilized without source data, so we attempt to recover and transfer the source domain knowledge by introduced data-free knowledge distillation approaches~\cite{lopes2017data,chen2019data,micaelli2019zero,fang2020data,yin2020dreaming} that are originally for model compression.

In this work, we propose a novel source-free unsupervised domain adaptation framework for segmentation, namely SFDA. 
Our framework alternatively works in two stages: knowledge transfer and model adaptation. Due to unavailable source data and uncertain target pseudo-labels, recovering and preserving the source knowledge learned by a source model is vital during adaptation training. This is because the uncertain supervision information in target pseudo-labels will tend to deviate the target model from the working domain. 
As such, in the knowledge transfer stage, we leverage a generator to estimate the source domain (working domain) and synthesize fake samples similar to the real source data in distribution, which can be used to transfer the domain knowledge from a well-trained source model to a target model. 
The key to semantic segmentation networks lies in capturing contextual feature relationships. 
With this intuition, a dual attention distillation (DAD) mechanism is introduced to help the generator synthesize samples with meaningful semantic context, which is beneficial to efficient pixel-level domain knowledge transfer. 
Moreover, the source model could work well on partial target domain and predict correct labels.
Therefore we propose an entropy-based intra-domain patch-level self-supervision module (IPSM) to leverage the correctly segmented patches as self-supervision during the model adaptation stage.

Our main contributions can be summarized as follows:
\begin{itemize}
  \item We propose the novel SFDA framework that combines knowledge transfer and model adaptation without requiring any source data and target labels.
  To our best knowledge, this is the first attempt to address the problem of source-free UDA for semantic segmentation.
  
  \item A novel dual attention distillation mechanism is designed specifically for segmentation to transfer and retain the contextual information, and the intra-domain patch-level self-supervision module is introduced to exploit patch-level knowledge in target domain. 
  
  \item We demonstrate the effectiveness of our framework on synthetic-to-real and cross-city segmentation scenarios. In particular, it can even achieve competitive results with the state-of-the-art source-driven UDA approaches under the source-free setting. 
\end{itemize}

\section{Related Work}

\noindent\textbf{UDA for Semantic Segmentation. } 
Existing UDA methods for segmentation can be mainly divided into three categories. To reduce the cross-domain discrepancy, numerous UDA methods~\cite{hong2018conditional,tsai2019domain,vu2019advent,pan2020unsupervised} focus on distribution consistency by introducing adversarial learning. Inspired by image-to-image translation~\cite{isola2017image,CycleGAN2017}, a category of UDA methods has been proposed to generate target images conditioned on source data~\cite{hong2018conditional,hoffman2018cycada}. In addition, self-supervision with target pseudo-labels is a relatively simple but efficient approach~\cite{chen2019maxsquare,zou2018unsupervised}, but it requires source data for supervision. 
In summary, all the above UDA methods for segmentation assume that the densely-annotated source dataset is available during adaptation, ignoring the data privacy and inaccessibility issues in practice. To the best of our knowledge, we are the first to consider the source-free unsupervised domain adaptation issue for image segmentation.

\noindent\textbf{Knowledge Distillation (KD). }
Knowledge distillation is originally developed to transfer knowledge from a large teacher network to a compact student network~\cite{hinton2015distilling}.
Since then, a variety of KD methods has been presented for model compression~\cite{liu2019knowledge,changyong2019knowledge,zhang2019fast,mukherjee2020tinymbert}, domain adaptation~\cite{zhao2019multi,zhou2020domain}, and multi-modal learning~\cite{zhao2018through,garcia2018modality,dou2020unpaired}. 
More recently, data-free knowledge distillation~\cite{lopes2017data,chen2019data} has drawn surging attention, due to the inevitable data privacy issue. 
In~\cite{lopes2017data,nayak2019zero}, activation records are used to reconstruct training samples for training a compact student model. Analogously, Batch Normalization Statistics (BNS) stored in Batch Normalization (BN) layers can be used to reconstruct training samples~\cite{yin2020dreaming,haroush2020the} as well. Most of the data-free KD methods based on generative adversarial networks~\cite{chen2019data,yoo2019knowledge,micaelli2019zero,fang2020data,ye2020data}. They all focus on generating fake samples for transferring knowledge from teacher to student networks without original training data mainly on classification tasks. In this work, we extend the data-free knowledge distillation methods to segmentation and tackle the source-free domain adaptation challenge.

\section{Methodology}

\begin{figure}[htbp]
  \centering
  \includegraphics[width=\linewidth]{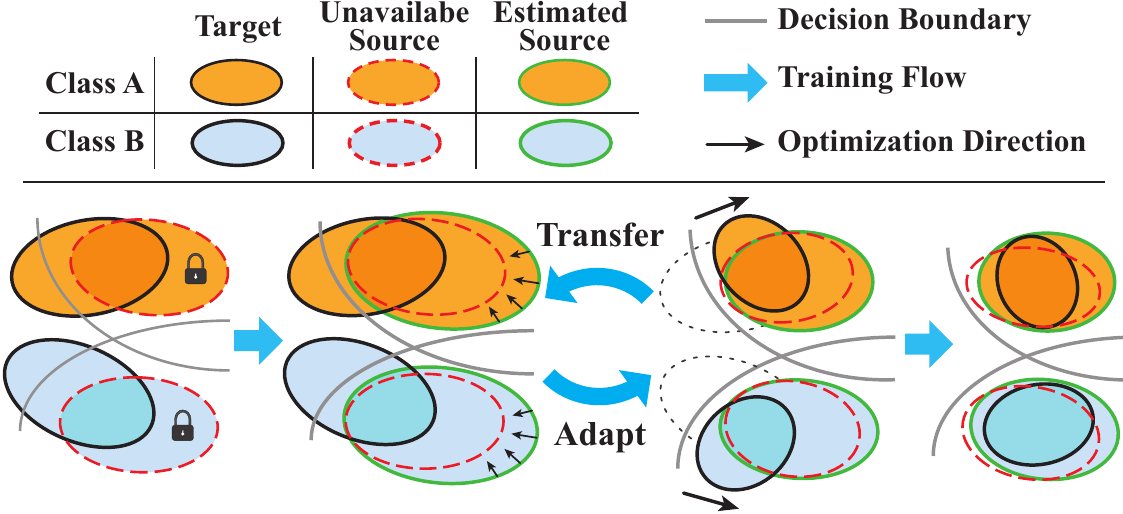}
  \caption{Overview of the training procedure for the proposed framework. Due to the unavailable source domain (marked as red dotted ellipse), we adopt a generator to estimate it by synthesizing fake samples (marked as green ellipse). } 
  \label{fig:principle}
\end{figure}

\subsection{Notations and Motivation}

For exiting source-driven UDA methods, an annotated source dataset $D_s = \{(x_s, y_s) | x_s\in \mathbb{R}^{H\times W \times 3}, y_s \in \mathbb{R}^{H\times W} \}$, an unlabeled target dataset $D_t = \{x_t | x_t \in \mathbb{R}^{H\times W \times 3}\}$ and a well-trained source model $\mathcal{S}$ are given. Note that $x_s$ and $x_t$ corresponds to the source and target sample, respectively, and $y_s$ is the label for the corresponding source image.
$H$ and $W$ are the height and width of the images.
The target model $\mathcal{T}$ generally shares parameters with the source model, but takes target data as input during adaptation. 
The source-driven UDA methods are commonly formulated by: 
\begin{equation}
  \mathcal{L}_{DA} = \mathcal{L}_{SEG}(x_s, y_s) + \mathcal{L}_{TAR}(x_t) \,,
\end{equation}
where $\mathcal{L}_{SEG}$ is the supervised training loss for preserving source domain knowledge, usually cross-entropy or focal loss. And $\mathcal{L}_{TAR}$ is the self-supervision loss for the target domain based on pseudo-labels, such as entropy minimization~\cite{vu2019advent}, maximum square loss (MaxSquare)~\cite{chen2019maxsquare}, \etc.
In this work, we adopt the maximum square loss as an assistance during adaptation, which is defined as:
\begin{equation}
  \mathcal{L}_{TAR}(x_t) = -\frac{1}{HW}\sum_{h,w}^{HW}\sum_{c}^{C}(p_{t}^{h,w,c})^2 \,,
\end{equation}
where $p_{t}^{h,w,c}$ is the probability of category $c$ for one target image pixel and $C$ is the number of semantic categories. 

In source-free scenarios, the annotated source dataset is unavailable, so the supervised learning process to preserving source knowledge will abort. Fortunately, the source domain knowledge has been permanently retained in the source model. 
We can consider source-free UDA as a knowledge transfer and adaptation problem, shown in Figure~\ref{fig:principle}. The orange or blue ellipse areas represent the feature space of the source and target domain. Due to the learning bias, the source model can only work well in the source domain, making it necessary to estimate the source domain (marked as green ellipse) and transfer the knowledge to target model during adaptation.  
Following the above principle analysis, a source-free UDA framework combining knowledge transfer and adaptation is proposed for semantic segmentation. 

\begin{figure}[!t]
  \centering
  \includegraphics[width=\linewidth]{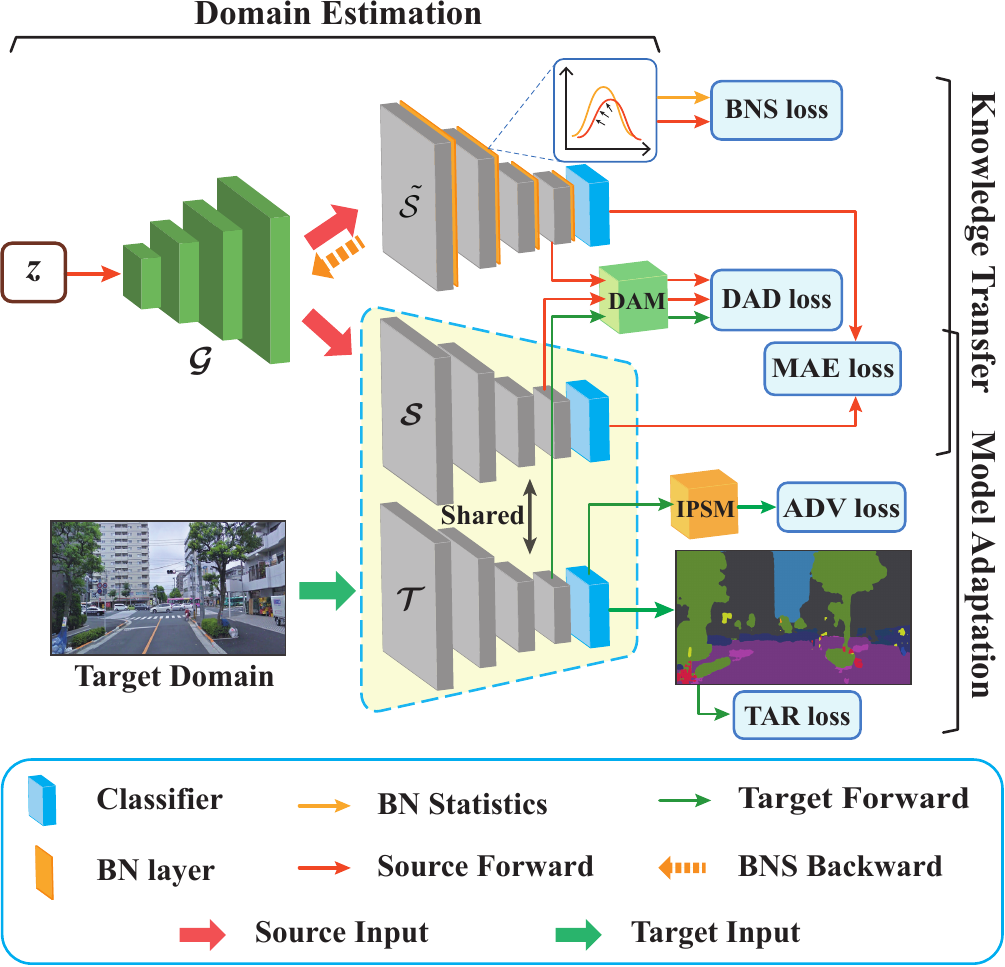}
  \caption{Architecture of the proposed SFDA framework.}
  \label{fig:framework}
\end{figure}

We denote the estimated source dataset with labels as $\tilde{D_{s}} = \{(\tilde{x}_s, \tilde{y}_s) | \tilde{x}_s\in \mathbb{R}^{H\times W \times 3}, \tilde{y}_s \in \mathbb{R}^{H\times W} \}$ (corresponding to the green ellipse in Figure~\ref{fig:principle}). Figure~\ref{fig:framework} shows our SFDA framework, which includes a \textbf{Knowledge Transfer} stage and a \textbf{Model Adaptation} stage. Note that, to preserve and transfer the source domain knowledge retained in the source model, we need to copy a source model $\tilde{\mathcal{S}}$ and fix its parameters in training. In the transfer stage, generator $\mathcal{G}$ synthesizes fake samples for transferring the source knowledge from the fixed source model $\tilde{\mathcal{S}}$ to $\mathcal{S}$. 
Moreover, an intra-domain patch-level self-supervision module (IPSM) is introduced to take advantage of information in patch-level pseudo-labels and improve the utilization of target data. We detail the two-stage SFDA in the following. 

\subsection{Source-Free Domain Knowledge Transfer}

\subsubsection{Source Domain Estimation}

To estimate the unavailable source domain, a generator $\mathcal{G}$ is designed to generate fake samples $\tilde{x}_s$ with random noises $z$ as input, drawn from a Gaussian distribution.
\begin{equation}
  \tilde{x}_s = \mathcal{G}(z), \; z \sim \mathcal{N}(\mathbf{0}, \mathbf{1}) \,.
\end{equation} 

Following BNS-guided data-free knowledge distillation~\cite{yin2020dreaming}, the feature distribution of estimated source samples is supposed to satisfy the batch normalization statistics of the source segmentation model. Hence, we apply a BNS constraint on the generator: 
\begin{equation}
  \mathcal{L}_{BNS} = \sum_{l}\| \mu_{l}(\tilde{x}_s) - \bar{\mu}_{l} \|^2_2 + 
  \sum_{l}\| \sigma_{l}^{2}(\tilde{x}_s) - \bar{\sigma}_{l}^{2} \|^2_2 \,,
\end{equation}
where $\tilde{x}_s$ is the synthetic data from the generator, $\mu_{l}(\tilde{x}_s)$ and $\sigma_{l}^{2}(\tilde{x}_s)$ are the batch-wise mean and variance estimates of feature maps at the $l$-th layer, and $\bar{\mu}_{l}$ and $\bar{\sigma}_{l}^{2}$ are the corresponding mean and variance parameters of the source domain stored in the $l$-th BN layer of source model $\tilde{\mathcal{S}}$. 

Different from~\cite{yin2020dreaming}, the generative approach for obtaining fake samples in our framework is more efficient and flexible, which avoids the time-consuming noise optimization procedure thanks to the generative adversarial knowledge transfer mechanism.
Specifically, for segmentation tasks, we construct a semantic-aware adversarial knowledge transfer mechanism, working based on the discrepancy between the source and target models.
To achieve this, we first formulate three different discrepancy measures for three models. The output space discrepancy between the fixed model $\tilde{\mathcal{S}}$ and the shared source model $\mathcal{S}$ is formulated as a mean absolute error (MAE): 
\begin{equation}
  \mathcal{L}_{MAE} = \mathbb{E}_{\tilde{x}_s}\left(\frac{1}{C}\| \mathcal{S}(\tilde{x}_s) - \tilde{y}_s\|_1\right) \,,
\end{equation}
where $\tilde{y}_s=\tilde{\mathcal{S}}(\tilde{x}_s)$ and $\mathcal{S}(\tilde{x}_s)$ are the prediction outputs from $\tilde{\mathcal{S}}$ and $\mathcal{S}$ for synthetic data $\tilde{x}_s$, respectively.  

Moreover, semantic information or contextual relationships performs a significant effect on segmentation. 
So the contextual relationships captured by the source model are supposed to be preserved and transferred.
The discrepancy of the contextual relationships between $\tilde{\mathcal{S}}$ and $\mathcal{S}$ is calculated by a dual attention distillation loss, which is given by:
\begin{equation}
  \mathcal{L}_{DAD}^{ss} = \mathbb{E}_{\tilde{x}_s}\left(\frac{1}{M}\|\mathcal{A}(\tilde{\mathcal{F}}^s(\tilde{x}_s)) - \mathcal{A}(\mathcal{F}^s(\tilde{x}_s))\|_1\right) \,,
\end{equation}
where $\mathcal{A}(\cdot)$ is the dual attention module (DAM) to calculate the dual attention map of the corresponding features. $M$ is the size of the attention map. 
$\tilde{\mathcal{F}}^s(\tilde{x}_s)$ and $\mathcal{F}^s(\tilde{x}_s)$ are the backbone feature extractors of the segmentation models $\tilde{\mathcal{S}}$ and $\mathcal{S}$ with synthetic data $\tilde{x}_s$ as input. 

Analogously, we can define the discrepancy between the source and target models as follows:
\begin{equation}
\begin{aligned}
  \mathcal{L}_{DAD}^{st} & = \mathbb{E}_{\tilde{x}_s}\left[ D_{KL}\left(S(\tilde{\mathcal{F}}^s(\tilde{x}_s)), S(\mathcal{F}^t(x_t) )\right) \right] \\
  & + \mathbb{E}_{\tilde{x}_s}\left[ D_{KL}\left(R(\tilde{\mathcal{F}}^s(\tilde{x}_s)), R(\mathcal{F}^t(x_t) )\right) \right] \,,
\end{aligned}
\end{equation}
in which $\mathcal{F}^t(x_t)$ obtains the feature map extracted from the backbone of target model with the target data $x_t$ as input. $S$ and $R$ are the spatial and channel attention maps extracted from the feature maps, which will be defined at Sec~\ref{sec:322}.
The motivation behind this equation is that the data generated by the generator is not enough to restore the contextual relationships of the source data, due to the lack of necessary prior information. 
Fortunately, the unlabeled target data has a similar domain-agnostic semantic structure with the real source data to a certain extent. 
This provides valuable knowledge for the generator to synthesize fake images.
So we adopt Kullback-Leibler (KL) divergence to measure the distribution distance of the dual attention maps of fake source and target data, then minimize it in optimization. 

\subsubsection{Dual Attention Module}
\label{sec:322}
\begin{figure}[htbp]
  \centering
  \includegraphics[width=\linewidth]{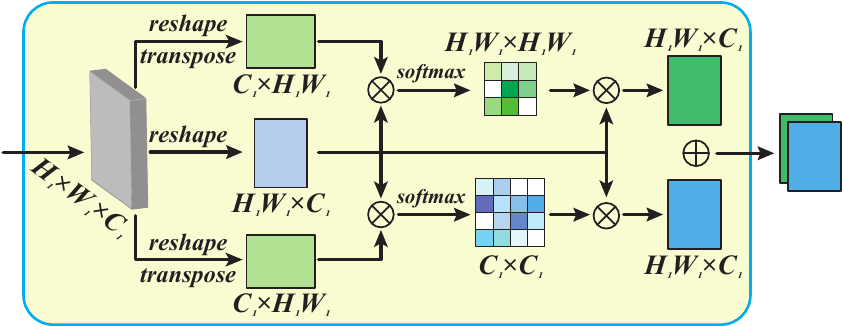}
  \caption{Dual attention module (DAM) for distillation. `$\otimes$' denotes matrix multiplication, and `$\oplus$' is a concatenation operator.} 
  \label{fig:DAM}
\end{figure}

In this section, we clarify the dual attention module.
The feature map extracted by backbone of segmentation network with $x$ as input is denoted as $F = \mathcal{F}(x), F \in \mathbb{R}^{H_1\times W_1\times C_1}$. Note that $H_1,W_1,C_1$ are the height, width and channel of the feature map respectively only in this sub-section. The dual attention module including spatial attention and channel attention is shown in Figure~\ref{fig:DAM}. Different from~\cite{fu2019dual,yang2020context}, we feed the feature $F$ into convolutional layers to generate new features, because DAM just aims to capture the spatial and channel-based long-range dependencies for distillation. 

To be specific, we first reshape $F$ so that $F\in \mathbb{R}^{N_1\times C_1}$, where $N_1 = H_1 \times W_1$ is the number of pixels.
$F^{\top}$ is the transpose of $F$.
Consequently, we calculate the spatial attention map $S \in \mathbb{R}^{N_1\times N_1}$ by:
\begin{equation}
  s_{ji} = \frac{\exp(F_{[i:]} \cdot F^{\top}_{[:j]})}{\sum^{N_1}_{i}\exp(F_{[i:]} \cdot F^{\top}_{[:j]})} \,,  
\end{equation}
where $s_{ji}$ measures the impact of the $i$-th position on the $j$-th position. 

Analogously, the channel attention map $R \in \mathbb{R}^{C_1\times C_1}$ can be calculated by:
\begin{equation}
  r_{ji} = \frac{\exp(F^{\top}_{[i:]} \cdot F_{[:j]})}{\sum^{C_1}_{i}\exp(F^{\top}_{[i:]} \cdot F_{[  :j]})} \,,  
\end{equation}
where $r_{ji}$ measures the impact of the $i$-th channel on the $j$-th channel. 

After obtaining the spatial and channel attention maps, the dual attention map of sample $x$ can be calculated by concatenating the two attention maps:

\begin{equation}
  \mathcal{A}(x) = \mathtt{concat}(F\cdot S | R \cdot F) \,.
\end{equation}
To transform the spatial and channel attention maps to the same shape, they are multiplied by the original feature $F$, respectively. 

\subsubsection{Objective Function}

In this way, we have introduced all the necessary components for source-free domain knowledge transfer (SFKT). The generator in our framework aims to synthesize valuable fake samples for transferring source knowledge from the source model to the target model. First, it is supposed to make the fake samples comply with the BNS constraints. Second, the generator explores the discrepancy space by maximizing the discrepancy between the source and target models to drive the search for new knowledge. In addition, it's better to take advantage of the prior attention information in the target domain by minimizing $\mathcal{L}^{st}_{DAD}$. Hence, the total objective function of generator is formulated as: 
\begin{equation}
  \min_{\mathcal{G}} {\mathcal{L}_{BNS} - \alpha\mathcal{L}_{MAE} - \beta\mathcal{L}_{DAD}^{ss} + \tau\mathcal{L}_{DAD}^{st}} \,,
\end{equation}
where $\alpha$, $\beta$ and $\tau$ are hyper-parameters for balancing the MAE loss and the two DAD losses. 

The target model learns from two aspects: the target pseudo-labels and two-level knowledge from the source model. We hope that while reducing the uncertainty of the target domain, the target also preserves the source domain information to guide adaptive learning by minimizing the output and attention discrepancy (two-level) with the source model. The objective function of target model in knowledge transfer stage is as follows:
\begin{equation}\label{eq:obj-stage1-TS}
  \min_{\mathcal{T},\mathcal{S}} {\alpha\mathcal{L}_{MAE} + \beta\mathcal{L}_{DAD}^{ss}} \,.
\end{equation}

\subsection{Self-supervised Model Adaptation} 

Since it is hard for the generator to guarantee to continuously restore and transfer the information precisely covering the source domain, we draw inspiration from the self-supervision mechanism and consider taking advantage of the valuable information output by target model for target data. 
Through analyzing the prediction of the initial target model on the target domain, we found that its prediction on most patches are correct, in which there are useful supervision information for learning on uncertain or error patches. 

To take advantage of the pseudo-labels in UDA-based segmentation, Pan \etal~\cite{pan2020unsupervised} proposed an unsupervised inter-domain and intra-domain adaptation method, which first separates the target domain into easy and hard splits using an entropy-based ranking function, and then decreases the inter-domain or intra-domain gap via an adversarial mechanism. However, in reality, the gap between the source and the target domain is too large, making it difficult to filter out a sufficient number of easy splits in the target domain for intra-domain supervision.
What makes matters worse is that the source domain is unavailable in our setting.

\begin{figure}[tbp]
  \centering
  \includegraphics[width=\linewidth]{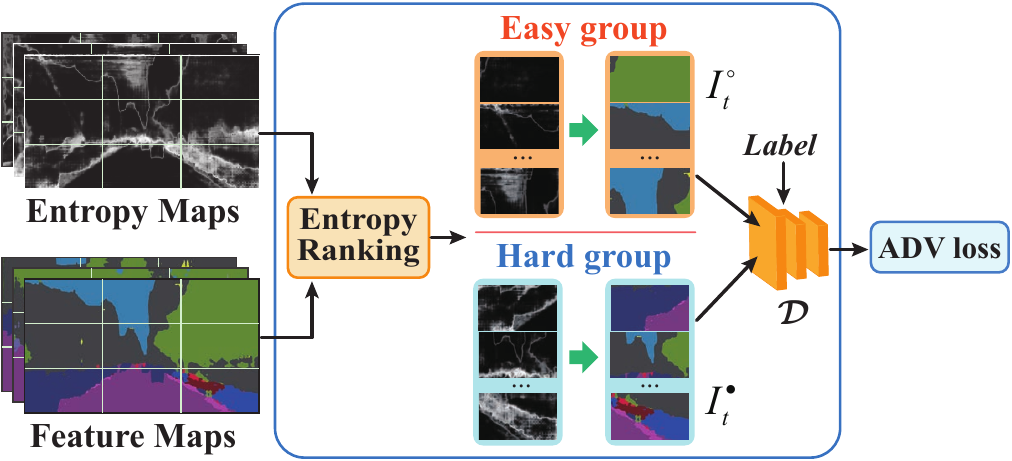}
  \caption{Intra-domain patch-level self-supervision module (IPSM).}
  \label{fig:IPSM}
\end{figure}

\begin{table*}[tbp]
  \centering
  \resizebox*{\linewidth}{!}{
    \begin{tabular}{c|c|c|ccccccccccccccccccc|c}
      \toprule
      Method & SF & \begin{sideways}Network\end{sideways} & \begin{turn}{45}road\end{turn} & \begin{turn}{45}sidewalk\end{turn} & \begin{turn}{45}building\end{turn} & \begin{turn}{45}wall\end{turn} & \begin{turn}{45}fence\end{turn} & \begin{turn}{45}pole\end{turn} & \begin{turn}{45}light\end{turn} & \begin{turn}{45}sign\end{turn} & \begin{turn}{45}veg.\end{turn} & \begin{turn}{45}terrain\end{turn} & \begin{turn}{45} sky\end{turn} & \begin{turn}{45}person\end{turn} & \begin{turn}{45}rider\end{turn} & \begin{turn}{45}car\end{turn} & \begin{turn}{45}truck\end{turn} & \begin{turn}{45}bus\end{turn} & \begin{turn}{45}train\end{turn} & \begin{turn}{45}motor\end{turn} & \begin{turn}{45}bike\end{turn} & mIoU \\
      \midrule
      source only & \xmark & \multirow{7}[2]{*}{\begin{sideways}DeepLabV3\end{sideways}} & 74.4  & 23.3  & 78.4  & 27.8  & 19.6  & 17.3  & 18.0  & 7.4   & 78.3  & 24.6  & 80.5  & 42.9  & 6.6   & 73.2  & 22.2  & 29.7  & 4.2   & 18.1  & 1.5   & 34.09 \\
      MinEnt~\cite{vu2019advent} & \xmark &      & 80.2  & 31.9  & 81.4  & 25.1  & 20.8  & 24.6  & 30.2  & 17.5  & 83.2  & 18.0  & 76.2  & 55.2  & 24.6  & 75.5  & 33.2  & 31.2  & 4.4   & 27.4  & 22.9  & 40.17 \\
      AdaptSegNet~\cite{tsai2018learning} & \xmark &      & 81.6  & 26.6  & 79.5  & 20.7  & 20.5  & 23.7  & 29.9  & \textbf{22.6} & 81.6  & 26.7  & 81.2  & 52.4  & 20.2  & 79.1  & \textbf{36.0} & 28.8  & \textbf{7.5} & 24.7  & 26.2  & 40.49 \\
      CBST~\cite{zou2018unsupervised}  & \xmark &      & 84.8  & \textbf{41.5} & 80.4  & 19.5  & \textbf{22.4} & 24.7  & 30.2  & 20.4  & \textbf{83.5} & 29.6  & 82.3  & 54.7  & 25.3  & 79.2  & 34.5  & 32.3  & 6.8   & 29.0  & \textbf{34.9} & 42.94 \\
      MaxSquare~\cite{chen2019maxsquare} & \xmark &      & \textbf{85.8} & 33.6  & 82.4  & 25.3  & 25.0  & \textbf{26.5} & \textbf{33.3} & 18.7  & 83.2  & \textbf{32.9} & 79.8  & 57.8  & 22.2  & \textbf{81.0} & 32.1  & 32.6  & 5.2   & \textbf{29.8} & 32.4  & 43.12 \\
      SFDA (w/o IPSM) & \cmark &      & 83.5  & 33.9  & 81.4  & 24.8  & 22.4  & 23.6  & 30.1  & 19.8  & 81.4  & 28.7  & 80.9  & 56.8  & 20.4  & 78.6  & 35.0  & 28.9  & 3.6   & 26.4  & 25.5  & 41.35 \\
      SFDA & \cmark &      & 84.2  & 39.2  & \textbf{82.7} & \textbf{27.5} & 22.1  & 25.9  & 31.1  & 21.9  & 82.4  & 30.5  & \textbf{85.3} & \textbf{58.7} & 22.1  & 80.0  & 33.1  & \textbf{31.5} & 3.6   & 27.8  & 30.6  & \textbf{43.16} \\
      \midrule
      source only & \xmark & \multirow{6}[2]{*}{\begin{sideways}SegNet\end{sideways}} & 48.9  & 17.2  & 76.4  & 6.7   & 12.5  & 22.8  & 12.6  & 4.8   & 77.2  & 15.1  & 74.2  & 47.2  & 7.2   & 57.7  & 20.3  & 10.2  & 1.0   & 2.2   & 1.1   & 27.13 \\
      MinEnt & \xmark &      & 79.8  & 31.7  & 78.8  & 20.2  & 18.4  & 23.9  & 14.7  & 4.9   & 80.6  & 17.9  & 78.4  & 48.9  & 5.2   & 77.6  & 21.7  & 17.1  & \textbf{12.7} & \textbf{10.5} & 2.6   & 33.97 \\
      AdaptSegNet & \xmark &       & 82.1  & 29.2  & 79.4  & 21.1  & 17.9  & 24.1  & 11.0  & \textbf{7.1} & \textbf{82.0} & 26.6  & 74.9  & 46.5  & 6.7   & 73.5  & 26.0  & 18.0  & 10.5  & 9.3   & 3.2   & 34.16 \\
      MaxSquare & \xmark &      & \textbf{82.9} & 33.6  & 80.2  & \textbf{22.7} & \textbf{20.2} & \textbf{26.3} & 15.5  & 6.1   & 81.8  & \textbf{27.5} & 78.8  & 48.3  & \textbf{10.1} & 79.8  & 24.4  & 20.1  & 13.2  & 9.4   & 5.3   & \textbf{36.11} \\
      SFDA (w/o IPSM) & \cmark &      & 80.5  & 30.3  & 81.6  & 24.5  & 18.0  & 25.1  & 13.7  & 3.2   & 79.4  & 25.6  & 76.3  & 44.6  & 7.3   & \textbf{80.5} & 24.7  & \textbf{21.4} & 10.5  & 4.4   & 2.5   & 34.43 \\
      SFDA & \cmark &      & 81.8  & \textbf{35.4} & \textbf{82.3} & 21.6  & 20.2  & 25.3  & \textbf{17.8} & 4.7   & 80.7  & 24.6  & \textbf{80.4} & \textbf{50.5} & 9.2   & 78.4  & \textbf{26.3} & 19.8  & 11.1  & 6.7   & 4.3   & 35.86 \\
      \bottomrule
    \end{tabular}%
  }
  \caption{Results on GTA5 $\rightarrow$ Cityscapes. `SF' represents whether the method is in source-free setting.}
  \label{tab:g2c}%
\end{table*}%

\subsubsection{Patch-level Self-supervision Module}

To cope with above issue, we present a novel entropy-based intra-domain patch-level self-supervision module to take advantage of the target domain pseudo-labels in the model adaptation stage, shown in Figure~\ref{fig:IPSM}. Considering in cityscapes segmentation scenarios, there are generally similar patterns or objects in the same areas of different street view images. Hence, we can leverage correct information at the patch level, which not only expands the samples but also alleviates uncertainty in entire pseudo-labels. In order to alleviate the difficulty of separating easy and hard samples caused by too large domain gap~\cite{pan2020unsupervised}, we divide each sample into $K\times K$ classes of sub-images or patches with a label $k$ ($k\in \{\mathbb{R}^{K^2}\}$) according to their positions. In prediction, the patches with lower entropy might have higher confidence and accuracy.
Hence the patches are split into easy and hard groups by entropy-ranking. 

We denote the height and width of each patch $x_{t,k}$ in target data $x_t \in \mathbb{R}^{H \times W \times C}$ as $H_2 = H/K$ and $W_2 = W/K$, and the corresponding prediction map output by the target model is $i_{t,k} \in\mathbb{R}^{H_2\times W_2\times C}$, $C$ is both the number of semantic categories and the channel of prediction maps. The probability map $p_{t,k}$ of patch $x_{t,k}$ can be calculated by a $\mathtt{softmax}$ function. 

Then, the mean entropy score of each prediction map $p_{t,k}$ for the target image $x_t$ is defined as:
\begin{equation}
  E(x_{t,k}) = -\frac{1}{H_{2}W_{2}}\sum^{H_{2}W_{2}}_{h,w}\sum^{C}_{c}p^{h,w,c}_{t,k} \log(p^{h,w,c}_{t,k}) \,.
\end{equation}

In a batch containing $B$ (even number) target images $\{x^b_{t,k} | b \in \{1,...,B\}\}$, entropy-ranking is executed on patch entropy maps at the same position or class. The $B/2$ prediction maps in each class with lower entropy are assigned to the easy group $I_{t,k}^{\circ} = \{i_{t,k}^{e} | e\in \{1,...,B/2\} \}$ , while the other $B/2$ are assigned to the hard group $I_{t,k}^{\bullet} = \{i_{t,k}^{d} | d\in\{1,...,B/2\} \}$. This process is given as follows:
\begin{equation}
  \begin{aligned} 
  I_{t,k}^{\bullet}, I_{t,k}^{\circ} \leftarrow \mathtt{Rank}(\{E(x_{t,k}^{b}) | b\in \mathbb{R}^{B} \}), \, k \in \mathbb{R}^{K^2} \,.
\end{aligned}
\end{equation}

After obtaining the prediction maps of hard and easy patches, we train a discriminator $\mathcal{D}$.
$\mathcal{D}$ aims to discriminates easy and hard patches, while $\mathcal{T}$ is trained to fool $\mathcal{D}$ from the side of hard patches to reduce the gap between patches. The adversarial learning loss to optimize $\mathcal{T}$ and $\mathcal{D}$ is given by:
\begin{equation}
  \begin{aligned}
  \mathcal{L}_{ADV}(I_{t}^{\bullet}, I_{t}^{\circ}) = -\sum^{K^2}_{k}\sum^{B/2}_{d,e} \log\left(1-\mathcal{D}(k, i_{t,k}^{e})\right) \\
  + \log\left(\mathcal{D}(k, i_{t,k}^{d})\right) \,.
  \end{aligned}
\end{equation}

\subsubsection{Objective Function}
Upon this, we extend the objective function in Equation~\ref{eq:obj-stage1-TS} by adding the adversarial loss \wrt IPSM and the self-supervision loss.
As a result, we define the following objective function to train the target and source models (\ie, $\mathcal{T}$ and $\mathcal{S}$) with shared weights:
\begin{equation}
  \min_{\mathcal{T},\mathcal{S}}\max_{\mathcal{D}} {\mathcal{L}_{TAR} + \alpha\mathcal{L}_{MAE} + \beta\mathcal{L}_{DAD}^{ss} + \gamma\mathcal{L}_{ADV}} \,,
\end{equation}
where $\gamma$ is the hyper-parameter to control the adversarial loss.
The detailed training algorithm is presented in the supplementary material.

\section{Experiments}

\subsection{Experimental Settings}

\subsubsection{Datasets and Metrics}
\textbf{Datasets } We evaluate our SFDA framework on semantic segmentation under two different settings:  synthetic-to-real and cross-city. 
For the former setting, we follow previous work~\cite{zou2018unsupervised,tsai2018learning} by considering Cityscapes~\cite{cordts2016cityscapes} as the target domain, and GTA5~\cite{richter2016playing} or SYNTHIA~\cite{ros2016synthia} as the source domain. 
For the latter setting, Cityscapes dataset is used as the source domain and NTHU~\cite{weng2016driver} dataset is as the target domain. 

Cityscapes~\cite{cordts2016cityscapes} provides 3,975 images with fine-grained segmentation annotations. The synthetic dataset GTA5~\cite{richter2016playing} contains 24,966 annotated images with a resolution of 1,914$\times$1,052 taken from the GTA5 game. SYNTHIA~\cite{ros2016synthia} is used as another synthetic dataset, which contains 9,400 fully annotated 1,280$\times$760 RGB images. The NTHU dataset~\cite{weng2016driver} contains four different cities: Rio, Rome, Tokyo, and Taipei.  

\noindent\textbf{Metrics } The semantic segmentation performance is evaluated on every category using Intersection-over-Union (IoU) ratio and Pixel-Accuracy (PA).
For the whole test set, we calculate Mean Intersection-over-Union (mIoU) and Mean Pixel-Accuracy (mPA).

\begin{figure*}[htbp]
  \centering
  \includegraphics[width=0.99\linewidth]{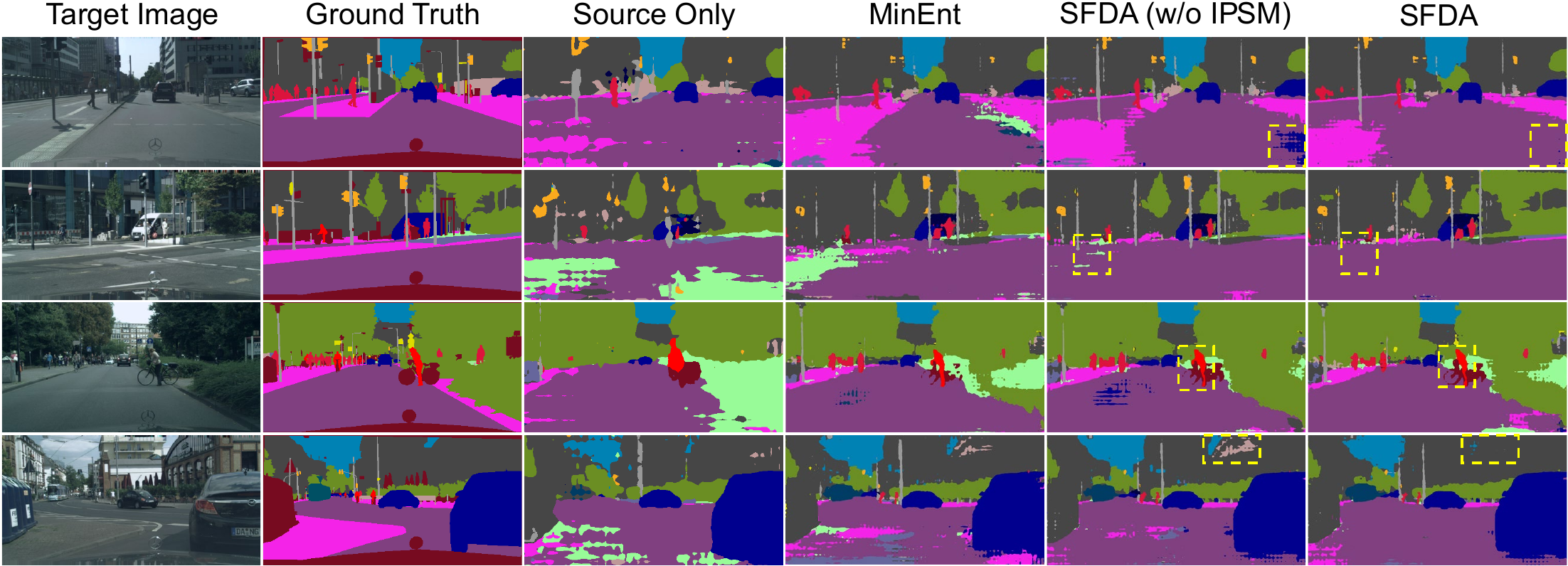}
  \caption{Qualitative results on GTA5 $\rightarrow$ Cityscapes. }
  \label{fig:g2c}
  \vspace{1mm}
\end{figure*}

\begin{table*}[htbp]
  \centering
  \resizebox*{\linewidth}{!}{
    \begin{tabular}{c|c|cccccccccccccccc|c|c}
      \toprule
      Method & SF & \begin{turn}{45}road\end{turn} & \begin{turn}{45}sidewalk\end{turn} & \begin{turn}{45}building\end{turn} & \begin{turn}{45}wall*\end{turn} & \begin{turn}{45}fence*\end{turn} & \begin{turn}{45}pole*\end{turn} & \begin{turn}{45}light\end{turn} & \begin{turn}{45}sign\end{turn} & \begin{turn}{45}veg.\end{turn} & \begin{turn}{45} sky\end{turn} & \begin{turn}{45}person\end{turn} & \begin{turn}{45}rider\end{turn} & \begin{turn}{45}car\end{turn} & \begin{turn}{45}bus\end{turn} & \begin{turn}{45}motor\end{turn} & \begin{turn}{45}bike\end{turn} & mIoU & mIoU* \\
      \midrule
      source only & \xmark & 55.5  & 21.0  & 75.8  & 2.2   & 0.1   & 19.9  & 0.2   & 6.9   & 74.5  & 81.5  & 37.1  & 8.7   & 45.1  & 18.2  & 2.5   & 19.8  & 29.31 & 34.36 \\
      MinEnt~\cite{vu2019advent} & \xmark & 78.2  & 39.6  & \textbf{81.9}  & 4.3   & 0.2   & \textbf{26.2} & 2.2   & 4.1   & 81.1  & 87.7  & 37.7  & 7.2   & 75.8  & 24.9  & 4.6   & 25.1  & 36.30 & 42.31 \\
      AdaptSegNet~\cite{tsai2018learning} & \xmark & 79.7  & 38.6  & 79.3  & 5.6   & \textbf{0.8} & 25.4  & 3.6   & 5.5   & 80.0  & 85.4  & \textbf{40.8} & 11.7  & 79.8  & 21.4  & 5.2   & 30.5  & 37.08 & 43.19 \\
      CBST~\cite{zou2018unsupervised} & \xmark & 81.4  & 44.2 & 80.4  & 7.9   & 0.7   & 25.6  & 5.2   & 12.4  & 81.4  & 89.5  & 39.7  & 10.6  & 82.1  & 21.9  & 6.3   & \textbf{32.9} & 38.88 & 45.23 \\
      MaxSquare~\cite{chen2019maxsquare} & \xmark & 81.0  & 39.8  & 82.6  & \textbf{8.7} & 0.5   & 23.2  & \textbf{6.6} & \textbf{12.4} & 85.3  & \textbf{90.1} & 39.9  & 8.4   & \textbf{84.7}  & 19.4  & \textbf{10.2} & 33.4  & 39.12 & 45.65 \\
      SFDA(w/o IPSM) & \cmark & 81.5 & 43.5  & 80.6  & 1.4   & 0.7   & 19.9  & 4.2   & 7.1   & 83.1  & 87.6  & 36.8  & 9.5   & 81.3  & 22.7  & 8.6   & 31.7  & 37.50 & 44.47 \\
      SFDA & \cmark & \textbf{81.9} & \textbf{44.9} & 81.7  & 4.0   & 0.5   & 26.2  & 3.3   & 10.7  & \textbf{86.3} & 89.4  & 37.9  & \textbf{13.4} & 80.6  & \textbf{25.6} & 9.6   & 31.3  & \textbf{39.20} & \textbf{45.89} \\
      \bottomrule
    \end{tabular}%
  }
  \caption{Results of domain adaptation task SYNTHIA $\rightarrow$ Cityscapes. `mIoU' and `mIoU*' are calculated over 16 and 13 classes, respectively.}
  \label{tab:s2c}%
\end{table*}%

\subsubsection{Implementation Details}
Two kinds of segmentation networks are adopted in our experiments. One is DeepLabV3~\cite{chen2017rethinking} with the ResNet-50~\cite{he2016deep} pre-trained on ImageNet~\cite{deng2009imagenet}, and the other is SegNet~\cite{badrinarayanan2017segnet} with the pre-trained VGG-16~\cite{simonyan2014very} backbone. Considering SegNet in an encoder-decoder architecture, the DAM is connected behind the encoder. When calculating the dual attention maps of target images, an adaptive pooling is applied before DAM. For the generator $\mathcal{G}$ and the discriminator $\mathcal{D}$, we use an architecture similar to~\cite{radford2015unsupervised} but extend $\mathcal{D}$ to a conditional version. The input channel of $\mathcal{D}$ is set to be consistent with the output channel of prediction maps. 
The latent space dimension for $\mathcal{G}$ and label embedding dimension for $\mathcal{D}$ both are 256. The architectures of the generator and the discriminator are detailed in the supplementary material. 

We implement the proposed framework using the PyTorch toolbox on two GTX 2080Ti GPUs. 
To train the segmentation networks, we use the Stochastic Gradient Descent (SGD) optimizer with Nesterov acceleration where the momentum is 0.9 and the weight decay is $10^{-4}$. The initial learning rate is set to $2.5 \times 10^{-4}$ and is decreased using the polynomial decay with a power of 0.9 as mentioned in~\cite{chen2017deeplab}.
For training the generator and discriminator, Adam optimizer~\cite{kingma2014adam} with an initial learning rate of 0.1 is adopted. 
Due to the difficulty in generating high-resolution images, we resize the images to 512$\times$256 for all datasets. Thanks to full-convolutional segmentation networks, we can set the resolution of synthetic samples to 256$\times$128, which is lower than target data but enough for transferring knowledge.  
To get a high-quality source model for adaptation, we pre-train the source models for 30 epochs on Cityscapes while for 20 epochs on GTA5 or SYNTHIA. In source-free adaptation, the target model, the generator, and the discriminator are jointly trained on a target domain for 120 epochs with a batch size of 8. 

As for hyper-parameters, $\alpha$ and $\beta$ are set to 1.0 and 0.5 by default, respectively. Notably we set $\tau=\beta$ to balance two DAD losses. We set $\gamma$ to 0.01 in all experiments if not particularly indicated. The number of patches, \ie, $K$ in IPSM is reasonable to choose from $\{3,4,5\}$. 

\subsection{Comparison}

\noindent\textbf{Synthetic-to-Real Adaptation:}
\textbf{(1) GTA5 $\rightarrow$ Cityscapes. }
Figure~\ref{fig:g2c} shows the qualitative results on GTA5 $\rightarrow$ Cityscapes. In order to show the versatility of SFKT and the contribution of IPSM, we remove the IPSM part in our architecture, namely `SFDA (w/o IPSM)'. It is obvious that even without source data, our method outperforms traditional MinEnt method. What's more, with the enhancement of IPSM, our full method can make up for errors in some areas through self-supervision, shown in the yellow dashed box. 
We present adaptation results in Table~\ref{tab:g2c} with comparisons to the state-of-the-art source-driven domain adaptation methods. 

\textbf{(2) SYNTHIA $\rightarrow$ Cityscapes. }
Following the evaluation setting in~\cite{vu2019advent,zou2018unsupervised}, we present the results of IoU and mIoU w.r.t. 16-class and 13-class segmentation in Table~\ref{tab:s2c}, respectively. Our architecture is used with DeepLabV3, and even outperforms the source-driven UDA methods with the assistance of IPSM. Besides, our method achieves competitive performance for the small object segmentation, such as traffic light, traffic sign, and motorbike. 

\noindent\textbf{Cross-City Adaptation:}
To show the effectiveness of our methods for smaller domain shift, we conduct the experiment on Cityscapes $\rightarrow$ NTHU with DeepLabV3 architecture. Table~\ref{tab:c2c} shows the comparisons of our method with other source-driven UDA methods.
Compared to the best UDA method MaxSquare, our method with IPSM achieves competitive performance on four city datasets.
In addition, we distill source domain knowledge via SFKT from well-trained source model into a new model and evaluate it on target domain without adaptation, shown as `transfer only' in the table. 
The results demonstrate that the knowledge we obtained via SFKT is still valuable on the target, although the effect is not as good as `source only'.

\begin{table}[htbp]
  \centering
  \resizebox*{\linewidth}{!}{
    \begin{tabular}{c|c|cccc}
      \toprule
      Method & SF & Rome  & Rio   & Tokyo & Taipei \\
      \midrule
      source only & \xmark & 46.44 & 45.06 & 44.05 & 44.07 \\
      MinEnt~\cite{vu2019advent} & \xmark & 47.29 & 46.82 & 45.49 & 45.12 \\
      AdaptSegNet~\cite{tsai2018learning} & \xmark & 47.99 & 47.81 & 46.22 & 45.13 \\
      MaxSquare~\cite{chen2019maxsquare} & \xmark & \textbf{48.48} & 48.74 & \textbf{47.10} & 47.16 \\
      \midrule
      transfer only & \cmark & 45.87 & 44.03	& 43.96 & 43.55 \\
      SFDA (w/o IPSM) & \cmark & 47.38 & 47.75 & 45.18 & 45.38 \\
      SFDA & \cmark & 48.33 & \textbf{49.03} & 46.36 & \textbf{47.20} \\
      \bottomrule
    \end{tabular}%
  }
  \caption{Results on Cross-City adaptation.}
  \label{tab:c2c}%
\end{table}%

\subsection{Ablation Study}

To show the detailed contributions of the components in SFKT, we conduct ablation experiments on three datasets, shown in Table~\ref{tab:abaltion_gen}. 
The results demonstrate that the DAD losses in source-free domain knowledge transfer is more effective than the commonly used BNS loss, and the fusion of them could further improve the performance.

\begin{table}[!t]
  \centering
  \resizebox*{\linewidth}{!}{
    \begin{tabular}{c|c|ccc}
    \toprule
    Dataset & source model & BNS   & DAD   & BNS+DAD \\
    \midrule
    GTA5~\cite{richter2016playing}  & 61.8  & 49.8  & 55.4  & 58.3 \\
    Cityscapes~\cite{cordts2016cityscapes}  & 73.6 &	60.6 & 65.5 & 70.8 \\
    SYNTHIA~\cite{ros2016synthia} & 62.3  & 51.4  & 54.7  & 59.0 \\
    \bottomrule
    \end{tabular}%
  }
  \caption{Results for key components in SFKT.}
  \label{tab:abaltion_gen}%
\end{table}%

\begin{figure}[htbp]
  \centering
  \includegraphics[width=\linewidth]{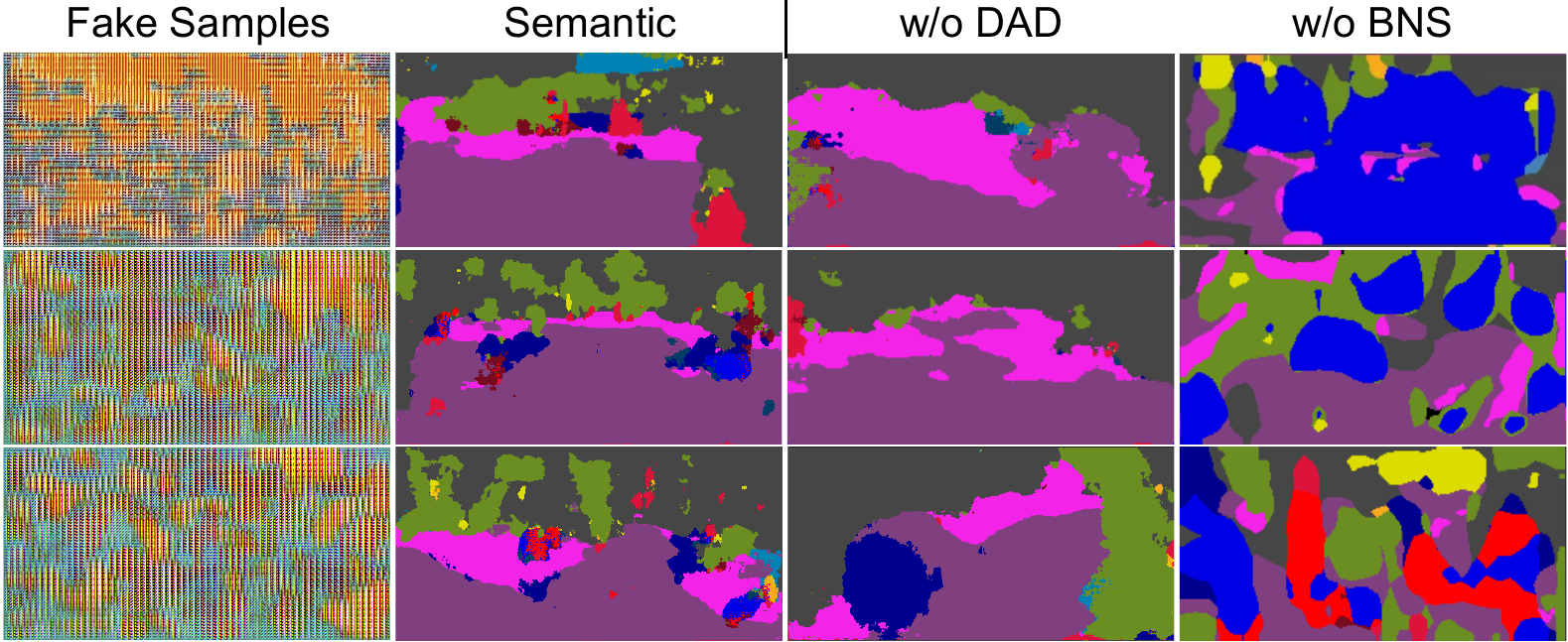}
  \caption{Visualization of synthetic semantic maps.}
  \label{fig:seg_gen}
\end{figure}

The visualization of semantic maps and fake samples synthesized in the knowledge transfer stage are shown in Figure~\ref{fig:seg_gen}. The left two columns are the fake samples synthesized by generator and corresponding semantic maps predicted by DeepLabV3 pre-trained on Cityscapes. The right two columns are several semantic maps predicted without DAD or BNS loss. On  one hand, the output semantic maps are similar to the real-world street view structure without DAD, but it is hard to pay attention to some small objects or refined segmentation. On the other hand, the generator captures the discrepancy between two models, but cannot preserve the original semantic distribution of source domain without BNS loss, which is vital for segmentation tasks. Although the fake samples cannot be recognized by humans, they have similar representations and outputs in convolutional neural networks with the source domain data. Hence, the fake samples become the key to transfer source domain knowledge. 

\subsection{Hyper-parameter Analysis}

Firstly, we discuss the influence of $\alpha$ and $\beta$ ($\tau=\beta$), the weights for the MAE loss and the DAD losses, respectively, for DeepLabV3 on GTA5 $\rightarrow$ Cityscapes. 
Given $\beta=0.5$, we adjust $\alpha$ from 0.1 to 2.0, and show the results in Table~\ref{tab:adjust_a}. Since the MAE loss $\mathcal{L}_{MAE}$ of the source prediction output is similar to the target segmentation loss $\mathcal{L}_{TAR}$ when supervised by target pseudo-labels, $\alpha$ should be close to 1.0. Otherwise, there will be disagreements with $\mathcal{L}_{MAE}$, resulting in bias during adaptation. 
\begin{table}[htbp]
  \centering
    \begin{tabular}{c|cccc}
    \toprule
    $\alpha$ & 0.1   & 0.5   & 1.0    & 2.0 \\
    \midrule
    mIoU  & 41.33 & 42.70  & 43.16 &  42.47 \\
    \bottomrule
    \end{tabular}%
    \caption{Influence of $\alpha$ given $\beta=0.5$.}
  \label{tab:adjust_a}%
\end{table}%

Analogously, given $\alpha=1.0$, we adjust $\beta$ from 0.01 to 1.0, and the results are shown in Table~\ref{tab:adjust_b}. Different from $\alpha$, $\beta$ controls the weights of the DAD losses in intermediate layers, so they are supposed to be smaller than $\alpha$. If too many weights are allocated to the DAD losses, they will limit the learning capacity of the intermediate layers. 
\begin{table}[htbp]
  \centering
    \begin{tabular}{c|cccc}
    \toprule
    $\beta$  & 0.01  & 0.1   & 0.5   & 1.0 \\
    \midrule
    mIoU  &  41.54 & 43.09 & 43.16 &  42.47 \\
    \bottomrule
    \end{tabular}%
    \caption{Influence of $\beta$ given $\alpha=1.0$.}
  \label{tab:adjust_b}%
\end{table}%

\begin{figure}[htbp]
  \centering
  \subfigure[GTA $\rightarrow$ Cityscapes]{
    \includegraphics[width=0.47\linewidth]{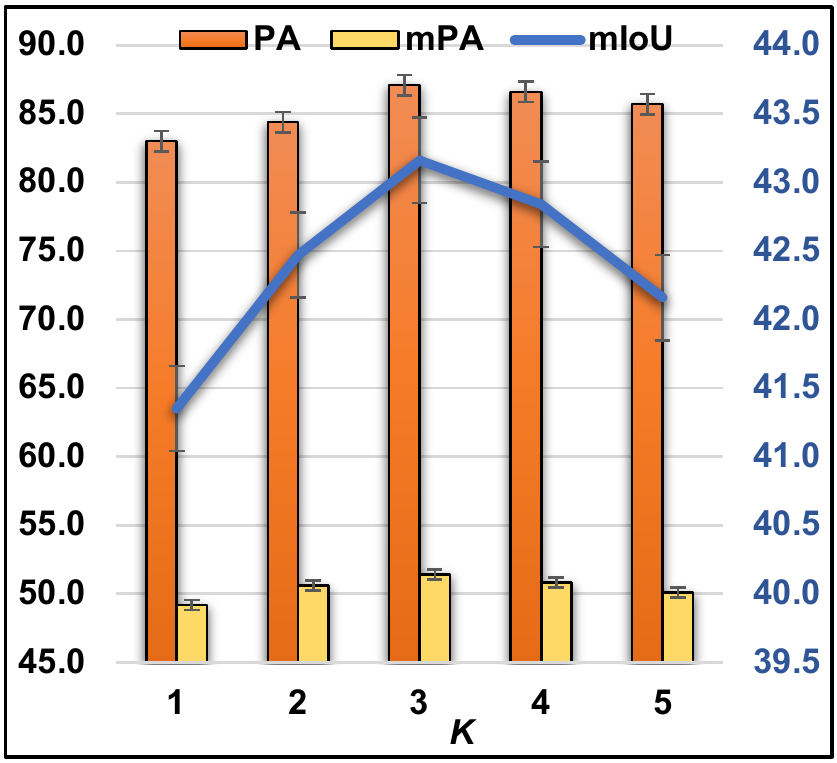}
  } 
  \subfigure[Cityscapes $\rightarrow$ Rio]{
    \includegraphics[width=0.47\linewidth]{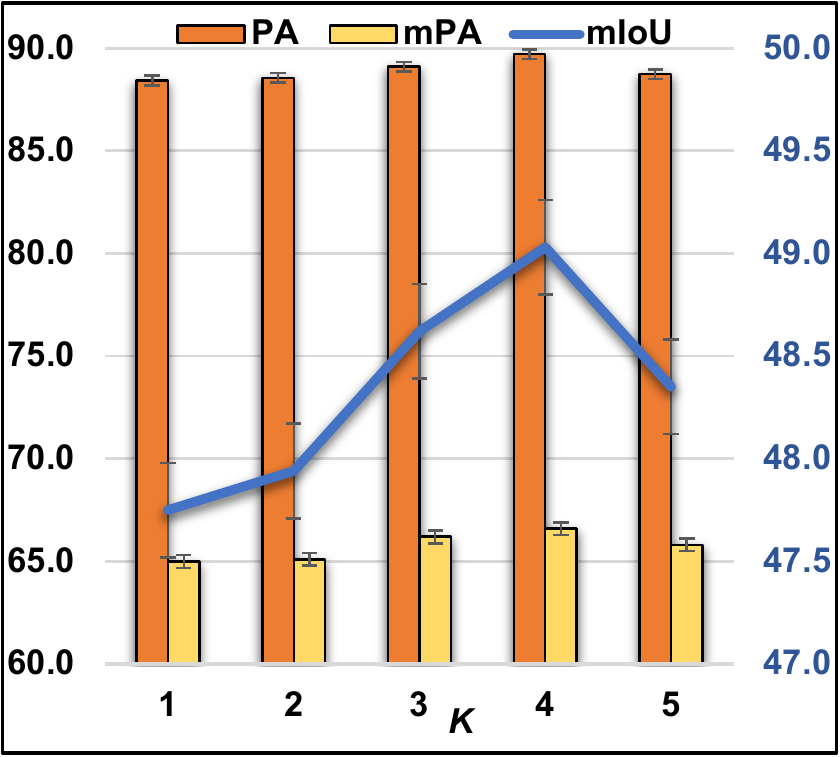}
  }
  \caption{Influence of number of patches (\ie, $K$).}
  \label{fig:ablation_K}
\end{figure}

We show the sensitivity analysis of parameters $K \in \{1,2,\cdots, 5\}$ in Figure~\ref{fig:ablation_K}, from which we observe that too large or too small $K$ is not suitable for IPSM, and 3 to 5 is reasonable. 
Note that when $K=1$, it means IPSM is not adopted in training. 

\section{Conclusion}

In this paper, we have presented a novel source-free domain adaptation framework (SFDA) for semantic segmentation. It aims to preserve the source domain knowledge from a fixed source model via knowledge transfer. 
Specifically, a dual attention distillation method is designed to capture and transfer pixel-level semantic information for segmentation tasks. Moreover, during model adaptation, an intra-domain patch-level self-supervision mechanism is introduced to take advantage of valuable knowledge at patch-level pseudo-labels in a target domain.
We conduct extensive experiments and ablation studies to validate the effectiveness of the proposed framework on different segmentation tasks, showing it performs favorably against existing source-driven UDA methods. However, our approach does not support high-resolution image segmentation tasks due to the limitation of generative fake sample synthesis, which will be tackled in future work.

\section*{Acknowledgement} 
This work was supported in part by National Natural Science Foundation of China under Grant (No. 62072182).

{\small
\bibliographystyle{ieee_fullname}
\bibliography{ref}
}

\end{document}